PAPER

# Development of a Deep Learning Model for the Prediction of Ventilator Weaning

Hernando González[1](✉),
Carlos Julio Arizmendi[1],
Beatriz F. Giraldo[2–4]

[1]Universidad Autónoma de Bucaramanga, Bucaramanga, Colombia

[2]Universitat Politècnica de Catalunya – Barcelona Tech (UPC), Barcelona, Spain

[3]Institute for Bioengineering of Catalonia (IBEC), The Barcelona Institute of Science and Technology, Barcelona, Spain

[4]CIBER de Bioingeniería, Biomateriales y Nanomedicina, Madrid, Spain

hgonzalez7@unab.edu.co

**ABSTRACT**

The issue of failed weaning is a critical concern in the intensive care unit (ICU) setting. This scenario occurs when a patient experiences difficulty maintaining spontaneous breathing and ensuring a patent airway within the first 48 hours after the withdrawal of mechanical ventilation. Approximately 20% of ICU patients experience this phenomenon, which has severe repercussions on their health. It also has a substantial impact on clinical evolution and mortality, which can increase by 25% to 50%. To address this issue, we propose a medical support system that uses a convolutional neural network (CNN) to assess a patient's suitability for disconnection from a mechanical ventilator after a spontaneous breathing test (SBT). During SBT, respiratory flow and electrocardiographic activity were recorded and after processed using time-frequency analysis (TFA) techniques. Two CNN architectures were evaluated in this study: one based on ResNet50, with parameters tuned using a Bayesian optimization algorithm, and another CNN designed from scratch, with its structure also adapted using a Bayesian optimization algorithm. The WEANDB database was used to train and evaluate both models. The results showed remarkable performance, with an average accuracy 98 ± 1.8% when using CNN from scratch. This model has significant implications for the ICU because it provides a reliable tool to enhance patient care by assisting clinicians in making timely and accurate decisions regarding weaning. This can potentially reduce the adverse outcomes associated with failed weaning events.

## 1    INTRODUCTION

Mechanical ventilation (MV) through an endotracheal tube, also known as invasive mechanical ventilation, is a frequently used intervention in patients admitted to the intensive care unit (ICU) [1]. It is a life-saving medical procedure used to assist or replace spontaneous breathing in patients with acute respiratory distress [2–3]. Studies have shown that approximately 40% of patients admitted to the ICU require invasive MV. The process of weaning patients from MV involves releasing them from mechanical support and removing the endotracheal tube (weaning). Assessing a patient's readiness for weaning from MV is a complex clinical task that varies internationally [4]. This process involves determining whether the patient's underlying disease has been successfully treated, assessing hemodynamic stability, evaluating the patient's level of consciousness, and reviewing current ventilator settings [5–6]. Automated weaning systems can improve the adaptation of mechanical support to the patient's ventilatory needs and facilitate systematic and early recognition of the patient's ability to breathe spontaneously and the possibility of discontinuing ventilation [7]. The Sixth International Conference on Intensive Care Medicine outlines readiness criteria for weaning patients from MV. Patients who meet the following readiness criteria satisfactorily should be considered ready for weaning: evidence of improvement in the underlying cause of MV, pH greater than 7.25, intact respiratory drive, hemodynamic stability with no or minimal cardiovascular support, tidal rate/volume ratio or rapid shallow breathing index (RSBI) less than 105 breaths/min/l, respiratory rate (RR) of 35 breaths/min or less, peak inspiratory pressure of –20 to –25 $cmH_2O$, vital capacity of greater than 10 ml/kg, and arterial oxygen saturation ($SaO_2$) greater than 90% with a $FiO_2$ of 0.4 or less (or an arterial oxygen partial pressure $PaO_2/FiO_2$ ratio of 150 mmHg or greater) [8]. Once these criteria are met, an assessment is performed using a diagnostic test called spontaneous breathing test (SBT), which is repeated every 24 hours to ensure successful extubating using unassisted T-breathing or low-level pressure support ventilation (PSV) [9]. SBT assists healthcare professionals in determining a patient's ability to maintain physiological breathing after weaning or during spontaneous breathing with a tracheostomy tube in place. Some patients do not meet weaning criteria within 20 minutes of SBT, so a 30-minute trial is recommended to assess the patient's ability to sustain spontaneous breathing. Although SBT is currently the preferred method for performing a weaning trial, it does not prevent the occurrence of complications after weaning, such as upper airway obstruction, increased resistance, coughing, and drainage of tracheobronchial secretions [10–12]. A well-performed SBT usually leads to successful weaning; however, failure of SBT requires a thorough investigation of potentially reversible conditions. A systematic and multidisciplinary approach is necessary to successfully deal with the weaning process. Prolonged weaning can be a waste of time and resources [13].

In the field of intensive care medicine, decision-making regarding the extubating of mechanically ventilated patients is a crucial challenge. Despite the application of well-established clinical criteria, the accuracy of predicting extubating outcomes remains unsatisfactory. This limitation has prompted the exploration of alternative approaches, particularly the use of machine learning (ML) techniques to exploit the information contained in the biomedical signals of the patient [14–15]. ML systems to predict extubating have gained significant attention in the medical community due to their ability to identify complex patterns and relate multiple variables that may influence the success or failure of the extubating process [16]. These systems use classification algorithms to establish robust predictive models that enable a more accurate assessment of the risk of reintubation and the need to prolong mechanical ventilation. To develop predictive models, biomedical data is collected, including physiological signals such as heart rate, oxygen saturation, and blood pressure obtained through continuous monitoring devices. Demographic data, clinical history, and laboratory test results can also be integrated to enrich the available information and improve the model's predictive capability. Several supervised learning techniques, including neural networks [17–19] and support vector machines [20], have been investigated to evaluate their ability to predict extubating outcomes in terms of sensitivity, specificity, and accuracy [21]. Rigorous validation of these models is crucial to ensuring their reliability and generalizability to different patient populations and clinical settings. Prospective and retrospective studies have been conducted using cohorts of patients with varying profiles and clinical characteristics to evaluate the performance of the proposed classifiers and their ability to improve clinical decision-making in the weaning setting.

Deep learning has emerged as a predominant method for analyzing large datasets and advancing artificial intelligence. The application of deep learning algorithms has led to significant breakthroughs in various domains, including face recognition, image processing, and speech recognition [22]. Within the realm of biomedical engineering, research has notably progressed with the adoption of classification methodologies leveraging deep learning models. Specifically, these classification systems utilize images of the power spectrum derived from time-frequency analyses of time-series data. This approach has been used in a wide range of clinical applications, including electrocardiography (ECG) [23–24], electromyographic (EMG) signals [25–26], encephalographic (EEG) signals [27–28], and extubating of mechanically ventilated patients [29]. Among the CNN architectures used, models such as VGG, ResNet, and DenseNet are notable for their ability to extract complex features from images and their effectiveness in classification tasks. CNNs can learn and extract complex features from spectral images, enabling more detailed and accurate analysis of these signals, which provides crucial information for clinical decision-making. The use of transforms, such as short-time Fourier transform (STFT), continuous wavelet transform (CWT), and nonlinear transforms, has enabled a more efficient and discriminative representation of the temporal and frequency characteristics of biomedical signals. This, in turn, has improved the ability of classification systems to distinguish between different physiological and pathological states. The integration of advanced signal analysis and ML techniques has expanded the potential for monitoring and diagnosing cardiovascular, neurological, and respiratory diseases while enhancing the precision and efficiency of clinical processes. CNNs are trained using two main approaches: transfer learning and training from scratch. Transfer learning involves leveraging the knowledge gained from a previously trained CNN on large and diverse datasets, such as ImageNet, and applying it to specific tasks in the biomedical field. This technique is particularly advantageous when dealing with limited datasets, as it enhances model performance with less training data. On the other hand, training from scratch entails initializing the CNN weights randomly and training the model using the biomedical domain-specific dataset. While this approach may necessitate more training data and computational time, it offers the benefit of fully customizing the neural network to the unique characteristics of biomedical signals and specific clinical applications. This customization can result in optimal performance on particular classification tasks.

The probability of premature weaning or prolongation of MV is high. The aim of this study is to introduce a deep learning model that can assist in the timely and accurate discontinuation of MV to aid physicians in decision-making. The study comprises three steps: first, processing the indices derived from time-frequency analysis (TFA) of the respiratory flow signal and the RR signal using wavelet transform; second, designing a classification system utilizing the power spectrum as an input image; and third, validating the proposed methodology.

## 2 MATERIALS AND METHODS

The respiratory flow and electrocardiographic signals utilized in this study demonstrate nonstationary behavior, indicating that their statistical properties change over time. This behavior can be attributed to the intricate interaction among the patient's underlying physiological processes during SBT. Employing TFA techniques, such as CWT, facilitates signal decomposition into different time and frequency scales. This approach enables a comprehensive examination of their evolution in both the time and frequency domains. Such methodology allows for the analysis of energy distribution in time series, significantly aiding in the comprehensive understanding of the underlying physiological processes and the identification of predictive patterns for diagnosis and clinical treatment.

### 2.1 Database

The Weandb database resulted from a study that included the electrocardiographic and respiratory flow signals of 133 patients who received MV and underwent extubating. These patients were registered in the Intensive Care Services of the Hospital de la Santa Creu i Sant Pau in Barcelona and the Hospital de Getafe, following approved ethical protocols. During the study, the patients underwent the T-tube test for 30 minutes (1800 s) of spontaneous breathing as part of the extubating protocol. The collected data can be used to develop pattern detection algorithms and predictive models for MV and extubating [30]. This information is valuable for developing algorithms and models for MV and extubating. Accoridng to the clinical criteria, after the SBT test, the patients were classified into three groups:

– The success group (Class $C_0$), includes 94 patients who successfully completed the weaning process.
– The failure group (Class $C_1$), consists of 39 patients who were unable to maintain spontaneous breathing and were reconnected to the ventilator after 30 minutes.
– The reintubated group (Class $C_2$), includes 21 patients who initially passed the 30-minute test but had to be reintubated and connected to a mechanical ventilator within 48 hours.

Eight time series were obtained through the ECG and the respiratory flow signals, for each patient: beat to beat interval obtained between each consecutive beat of the electrocardiographic signals (RR), interval between consecutive beats of an electrocardiographic signal, inspiratory time ($T_I$), expiratory time ($T_E$), breathing duration ($T_{Tot} = T_I + T_E$), tidal volume ($V_T$), inspiratory fraction ($T_I/T_{Tot}$), mean inspired flow ($V_T/T_I$), and frequency-tidal volume ratio ($f/V_T$), where $f$ is the respiratory rate. Figure 1 shows the information from the 8-time series of a patient in the success group. The time interval between samples is not uniform due to the nature of the parameters calculated from the electrocardiographic and respiratory flow signals, which vary in frequency and amplitude for each patient.

### 2.2 Data processing

Data preprocessing is a fundamental step in information analysis and decision-making, as raw data can be noisy, messy, and contain errors that can lead to incorrect conclusions. Database preprocessing involves performing basic tasks such as removing outliers, handling missing data, normalizing data, and coding variables. Outliers in the database are identified as data points that deviate from the mean by more than five standard deviations and are replaced by the mean of the three data points before and after the time point [31]. During an extubating test, patients may exhibit involuntary movements, making it challenging to obtain a continuous record of information during the 30-minute test. To address this issue, regions where continuous information is not observed are identified, and a threshold is set so that the distance between samples without information does not exceed 20 seconds. The longest continuous region from each dataset is then selected and stored in a variable. This process is repeated for each of the eight-time series recorded for each patient in classes $C_0$, $C_1$ and $C_2$. Z-score normalization is a statistical procedure that transforms a data series into a distribution with a mean of 0 and a standard deviation of 1 [32]. This normalization is beneficial because it allows for comparing data measured on different scales. To calculate the Z-score of a data series $x_n$, the following equation is used: where $\mu$ is the mean and $\sigma$ is the sample standard deviation.

$$Z_n = \frac{x_n - \mu}{\sigma} \qquad (1)$$

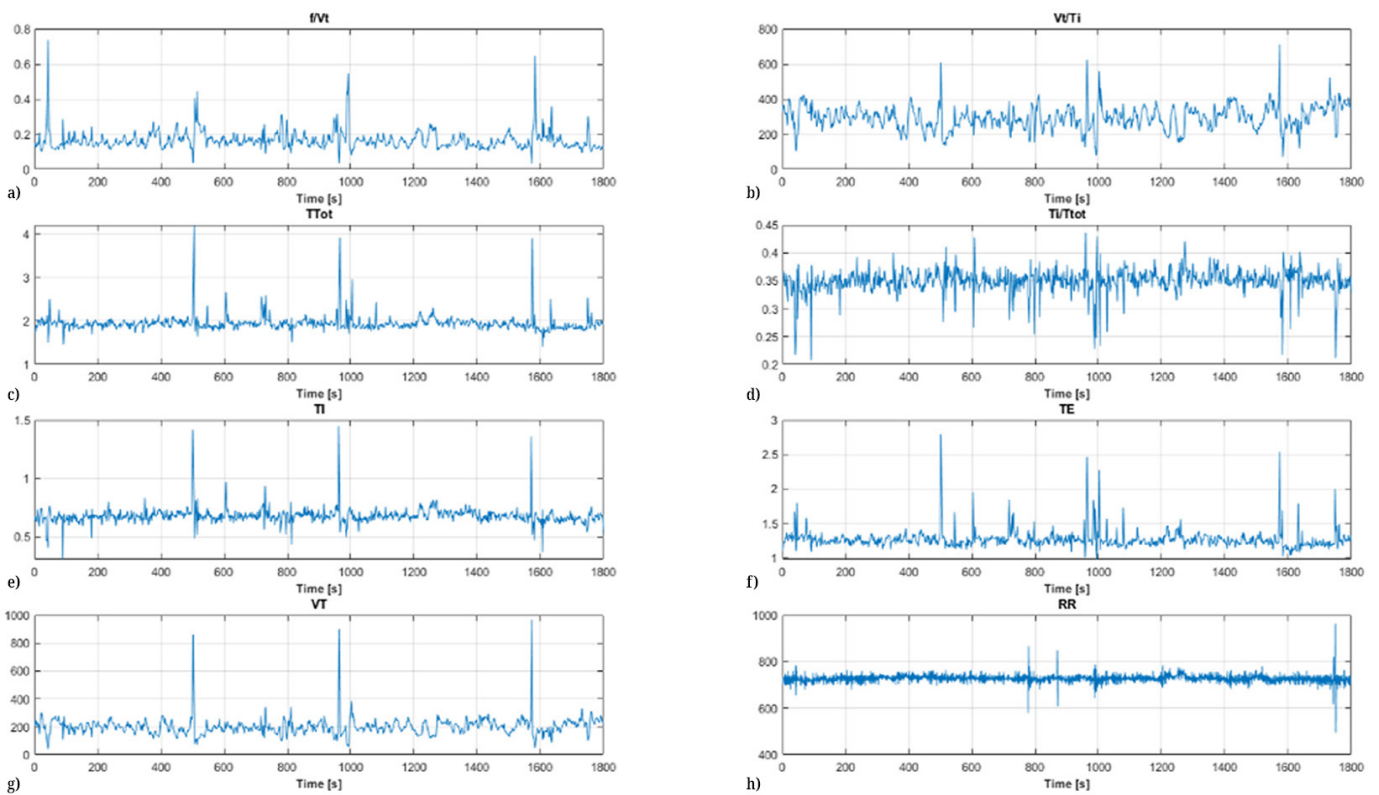

**Fig. 1.** Time series from the Weandb database. a) Frequency-tidal volume ratio $f/V_T$, b) Mean inspired flow $V_T/T_I$, c) Breathing duration $T_{Tot}$, d) Inspiratory fraction $T_I/T_{Tot}$, e) Inspiratory time $T_I$, f) Expiratory time $T_E$, g) Tidal volume $V_T$, h) Beat to beat interval $RR$

Several strategies exist for selecting the appropriate Nyquist frequency and resampling time for a finite set of unevenly sampled data. Existing methods are grouped into four main categories: least-squares-based methods, interpolation techniques, slotted resampling, and continuous-time models. In this study, to determine the interpolation frequency and interpolation method, the correlation index between the magnitude of the spectrum of the interpolated signal calculated using the discrete Fourier transform (DFT) and the magnitude of the spectrum of the original signal calculated using the non-uniform discrete Fourier transform (NUDFT) is calculated. The NUDFT for a data sequence $x_n$ is defined as

$$X(\omega_k) = \sum_{n=0}^{N-1} x_n e^{-j\omega_k t_n} \qquad (2)$$

where $X(\omega_k)$ is the NUDFT of the sequence $x_n$ evaluated at the angular frequency $\omega_k$, $N$ is the length of the data vector, and $t_n$ is the non-uniform time vector associated with the $x_n$ samples. If $\omega_k = \frac{2\pi k}{NT}$, where $T$ is the sampling time and $k = 0, 1, 2, ..., N-1$, a DFT-like expression is obtained [33]. The correlation between two signals, in the context of signal processing, measures the similarity or linear relationship between two data sets. The general expression to calculate the correlation between two discrete signals $x[n]$ and $y[n]$ of length $N$ is:

$$C(x,y) = \frac{\sum_{n=0}^{N-1}(x[n]-\bar{x})(y[n]-\bar{y})}{\sqrt{\sum_{n=0}^{N-1}(x[n]-\bar{x})^2 \sum_{n=0}^{N-1}(y[n]-\bar{y})^2}} \qquad (3)$$

where $\bar{x}$ and $\bar{y}$ are the mean values of $x[n]$ and $y[n]$, respectively. A value close to zero indicates that there is no linear relationship between the two signals, and a correlation of ±1 indicates a linear relationship between the two signals. Seven interpolation methods have been evaluated: *linear*, which performs a linear interpolation between two consecutive data points; *nearest*, which assigns the value of the point closest to the query point; *next*, which assigns the value of the next data point after the query point; *previous*, which assigns the value of the data point before the query point; piecewise cubic Hermite interpolating polynomial (pchip), which is an interpolation that uses cubic Hermite polynomials to interpolate data into local segments [34]; *spline*, which uses cubic polynomials to interpolate between data points; and modified Akima piecewise cubic Hermite interpolating polynomial (makima), which combines the pchip and spline methods to reduce oscillations and produce smoother results [35]. For each interpolation method, an iteration is performed on a vector of frequencies from 0.1 Hz to 3 Hz, corresponding to a sampling period between 0.33 s and 10 s, with increments of 0.1 Hz; this frequency range covers the range of separation between samples of the data recorded in the Weandb database. Figure 2 shows the mean behavior of the correlation index for each combination of the interpolation method and the interpolation frequency for the 134 patients comprising the database, class $C_0$, $C_1$ and $C_2$. The optimal frequency for the interpolation of the $T_I$, $T_E$, $T_{Tot}$, $V_T$, $T_I/T_{Tot}$, $V_T/T_I$ and $f/V_T$ signals is 0.3 Hz with the linear interpolation method, for the *RR* signal, the interpolation frequency is 1.5 Hz with the pchip method; this procedure ensures an interpolation that preserves the spectral information adequately, guaranteeing accuracy in the subsequent analysis of the signal. A dispersion analysis of the data is performed for the seven methods, using the appropriate interpolation frequencies for each signal. The results are presented in a box plot (Figure 3). This plot provides a visual representation of the distribution of the data and shows that there are few outliers for each interpolation method.

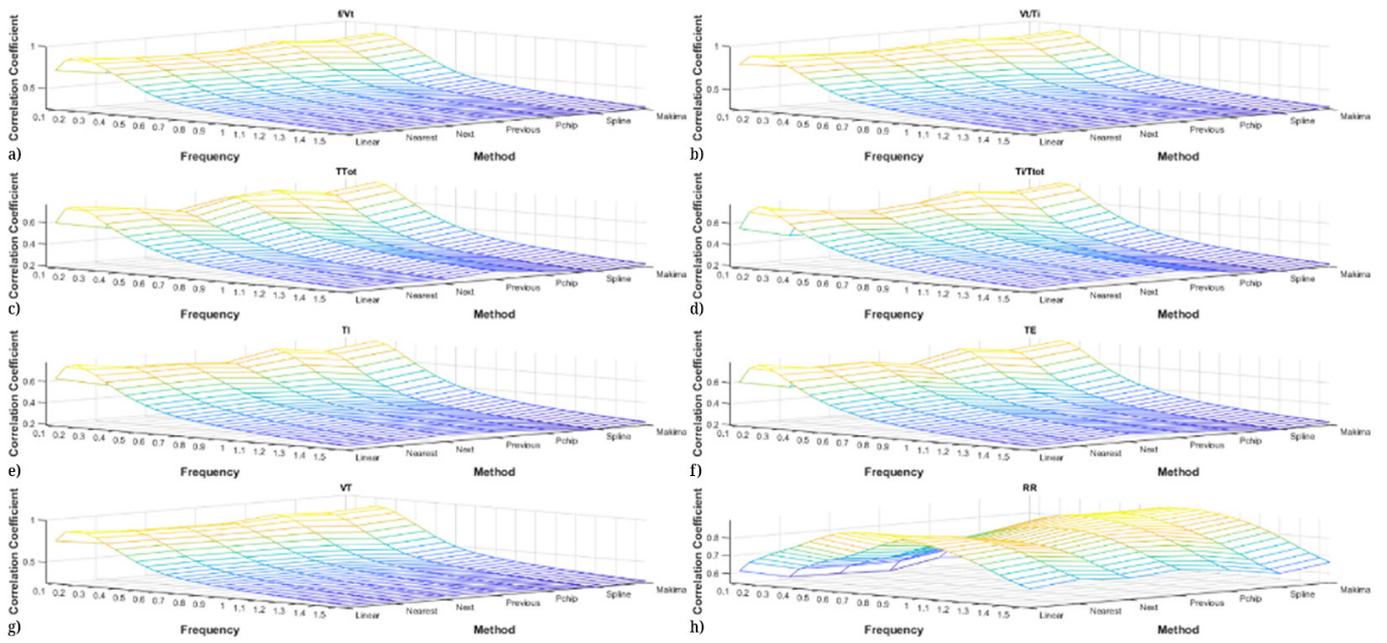

**Fig. 2.** Mean cross-correlation index as a function of interpolation frequency and interpolation method. a) Frequency-tidal volume ratio $f/V_T$, b) Mean inspired flow $V_T/T_I$, c) Breathing duration $T_{Tot}$, d) Inspiratory fraction $T_I/T_{Tot}$, e) Inspiratory time $T_I$, f) Expiratory time $T_E$, g) Tidal volume $V_T$, h) Beat to beat interval $RR$

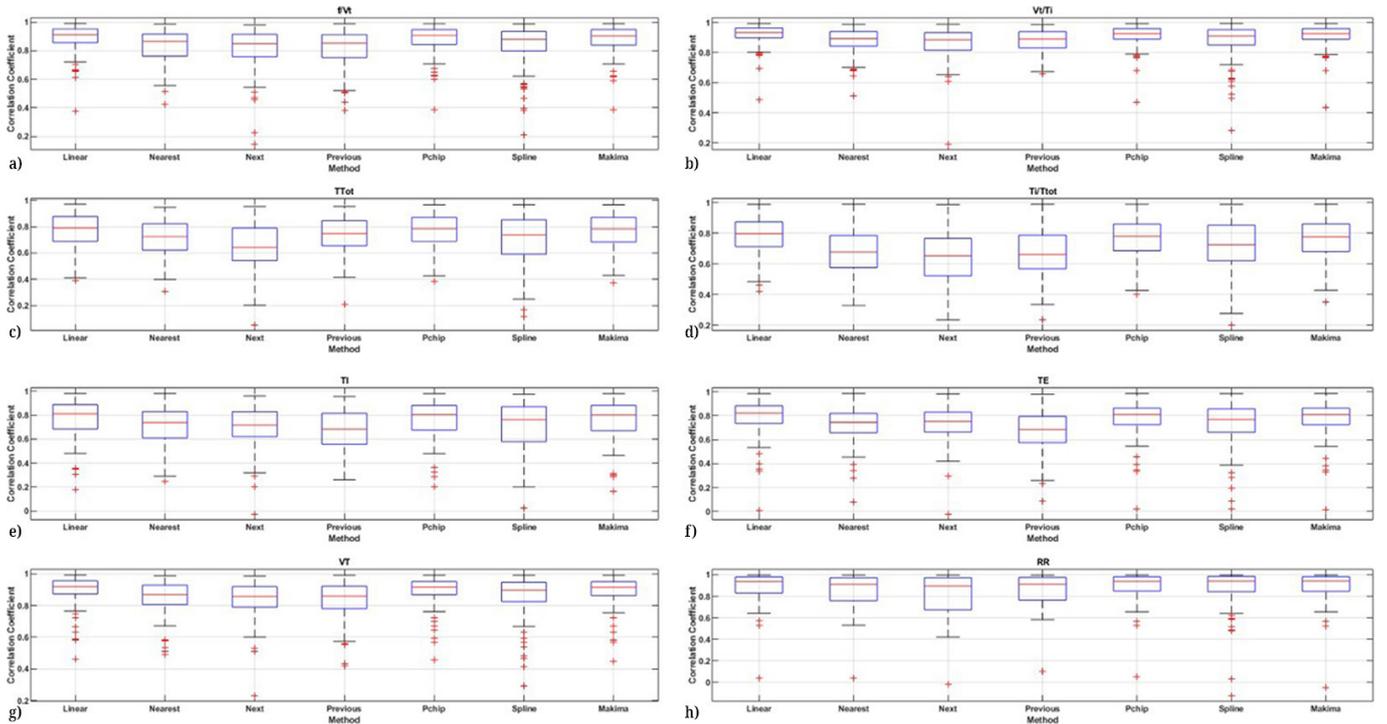

**Fig. 3.** Boxplot for a constant interpolation frequency. a) Frequency-tidal volume ratio $f/V_T$, b) Mean inspired flow $V_T/T_I$, c) Breathing duration $T_{Tot}$, d) Inspiratory fraction $T_I/T_{Tot}$, e) Inspiratory time $T_I$, f) Expiratory time $T_E$, g) Tidal volume $V_T$, h) Beat to beat interval $RR$

### 2.3 Time-frequency analysis

CWT is a signal analysis technique that facilitates the investigation of frequency variability in a data sequence over time. Unlike DFT, which provides information about the frequencies present in a signal but not their temporal location, CWT highlights both the frequency and temporal location of signal features. CWT is defined as:

$$S(a,b) = \frac{1}{\sqrt{a}} \int_{-\infty}^{\infty} x(t)\psi\left(\frac{t-b}{a}\right)dt \qquad (4)$$

where $x(t)$ is the input signal, $\psi$ is the mother wavelet function, $a$ is the scaling factor that controls the wavelet width, and $b$ is the translation parameter that controls the location of the wavelet. By varying the wavelet scale factor, CWT can capture both high-frequency features, using small-scale wavelets, and low-frequency features, using large-scale wavelets, present in a signal. This adaptive capability positions it as a tool for the analysis of signals containing transients, interruptions, discontinuities, and other short-duration phenomena typically found in non-stationary signals. The power spectral density (PSD) of the eight time series is evaluated by defining the Morse function and the Morlet function as the mother wavelet function. The two wavelet functions are designed to represent information in both the time and frequency domains of a signal and differ in their frequency profiles. The Morse function has a more flexible and adjustable frequency profile, allowing it to capture a wide range of signal characteristics, including short-duration transients and abrupt frequency changes [36]. On the other hand, the Morlet function is more closely related to the Gaussian function and has a constant center frequency with a fixed bandwidth, making it more suitable for analyzing signals that have more stationary and smooth frequency structures [37]. The methodology for selecting the mother wavelet function involves evaluating the normalized cross-correlation between the power spectra of patients in class $C_0$ and those in class $C_1$ for the eight-time series in the database. The correlation matrix is then obtained, and its mean and variance are calculated. The selection criterion is based on identifying the mother wavelet function that results in average correlations close to zero and minimum variance. An average correlation close to zero indicates that the parent wavelet function does not introduce biases in the correlations between success and failure signals, suggesting a greater ability to detect significant patterns in the data. On the other hand, a minimum variance indicates a reduced dispersion of correlation values, reflecting greater consistency in the obtained correlations. Figure 4c shows the normalized cross-correlation between two images: one associated with the PSD of the $f/V_T$ time series of a $C_0$ patient (Figure 4a) and the PSD of a $C_1$ patient for the same signal. The graph illustrates that the magnitude varies between –0.3 and 0.3, indicating the dissimilarity between the two images. Therefore, a CNN algorithm can be used to identify differences and design a classifier to determine if a patient can be extubated after an SBT process. Table 1 shows the results of the mean and average variance of the normalized cross-correlation matrices for the two mother wavelet functions. The Morse wavelet obtains the most favorable values in terms of mean and variance when analyzing between success group patients and failure group patients. These results suggest that the Morse wavelet is able to more accurately capture the patterns present in the database signals.

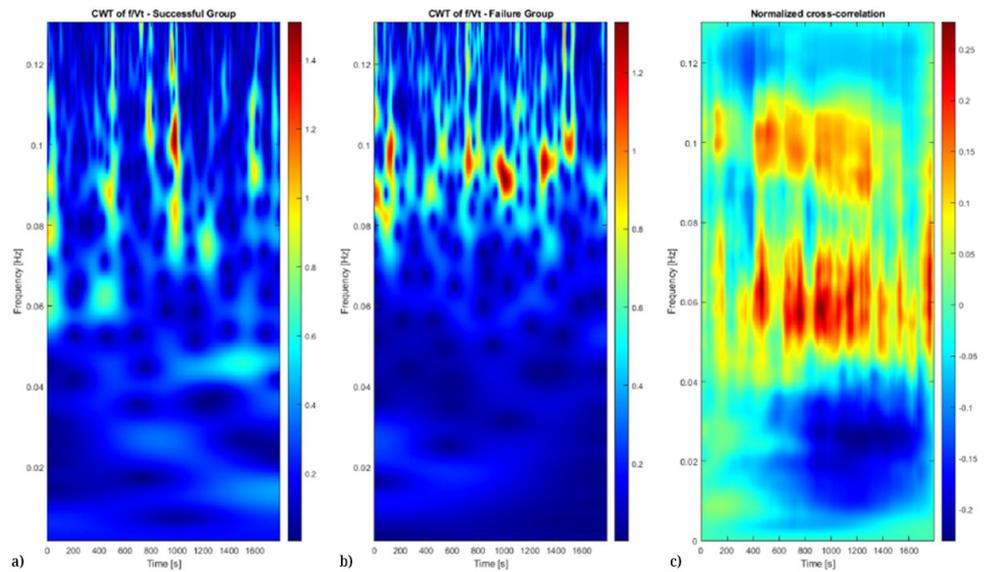

**Fig. 4.** Cross-correlation analysis. a) PSD of the $f/V_T$ signal of a patient class $C_1$. b) PSD of the $f/V_T$ signal of a patient class $C_0$. c) Normalized cross-correlation of a success group patient with a failure group patient

Table 1. Average of the mean and variance of the normalized cross-correlation matrix for success group patients and failure group patients

| Variable | Morse Wavelet | | Morlet Wavelet | |
|---|---|---|---|---|
| | Mean | Variance | Media | Variance |
| $f/V_T$ | 0.1139 | 0.0068 | 0.1181 | 0.0073 |
| $V_T/T_I$ | 0.1095 | 0.0070 | 0.1134 | 0.0074 |
| $T_{Tot}$ | 0.1237 | 0.0077 | 0.1271 | 0.0081 |
| $T_I/T_{Tot}$ | 0.1467 | 0.0098 | 0.1488 | 0.0102 |
| $T_I$ | 0.1241 | 0.0080 | 0.1257 | 0.0082 |
| $T_E$ | 0.1333 | 0.0084 | 0.1372 | 0.0088 |
| $V_T$ | 0.1152 | 0.0075 | 0.1199 | 0.0080 |
| RR | 0.1396 | 0.0104 | 0.1400 | 0.0105 |

### 2.4 Classification system

Computer vision is an interdisciplinary field that combines computer science, artificial intelligence, and visual perception to solve complex problems related to image interpretation and understanding. CNNs have proven to be essential tools for designing efficient and accurate classifiers. These networks are inspired by the functioning of the human visual system and can automatically learn relevant image features through training on labeled datasets. The objective of this work is to create two classifiers that can determine whether a patient can be disconnected from a VM after SBT. The first classifier is a ResNet-50 CNN [38], and the second is a CNN developed from scratch. The input images for the classifiers correspond to the PSD obtained with the CWT, with a resolution of 224 × 224 pixels.

For the ResNet50 CNN architecture, one CNN is trained for each of the eight-time series: $T_I$, $T_E$, $T_{Tot}$, $V_T$, $T_I/T_{Tot}$, $V_T/T_I$, $f/V_T$, and RR replaces the final classification layer with a global averaging layer. This is followed by a fully connected layer with a ReLU activation function and an output layer with a softmax activation function. Hyperparameter tuning is a crucial step in obtaining an optimal solution. The hyperparameters can be divided into two categories: those related to training, such as learning rate, batch size, dropout rate, and epoch count, and those related to model design, such as model structure, regularizes, and activation functions [39]. To optimize the performance of each ResNet50 CNN, the number of neurons in the fully connected layer is adjusted from 32 to 512 neurons, with a step of 32 neurons. Additionally, the appropriate optimizer (Adam or SGD), learning rate, and batch size are selected using the Keras Tuner Toolkit Bayesian optimization algorithm. Bayesian optimization offers advantages over other methods, such as manual search, grid search, and random search [40–41].

– The algorithm utilizes previous iterations to approach the perfect solution by selecting and testing alternative hyperparameters.
– It requires less time to converge to an optimal solution compared to a grid search because it does not test all possible combinations of hyperparameters.
– When dealing with large amounts of data and computational density, such as in training deep learning models, this tool can be particularly helpful.

In the hyperparameter optimization process, an early stopping technique is used to prevent overfitting and reduce computational time. This technique interrupts training when a specific metric, in this case the validation loss, stops improving for a predefined number of epochs. To establish a single output value for the classifier, it uses a method called receiver operating characteristic (ROC)-AUC weighted prediction. This method multiplies the probability associated with each classifier by its respective area under the ROC curve. This allows for the integration of information from multiple models into a single classifier. The ROC curve is generated using validation data by plotting the true positive rate (TPR) against the false positive rate (FPR) at different classification thresholds.

The CNN is built from scratch with an input size of 224 × 224 × 8, where each channel represents the PSD of the respiratory flow and electrocardiographic signals obtained through the CWT. The CNN architecture is determined through an exhaustive search of hyperparameters using the Bayesian optimization algorithm. This process determines the number of convolutional layers, the number of filters in each convolutional layer, the number of neurons in the fully connected layers, and the size of the training batch. Various configurations were tested, ranging from 2 to 5 convolutional layers with different numbers of filters (32, 64, and 128). Additionally, models with 1 to 3 fully connected layers, each containing 256, 512, or 1024 neurons, were evaluated. The learning rate was fine-tuned within the range of 1e-6 to 1e-3, while the training batch size varied between 10 and 100, with a default value of 32. The model was compiled using the Adam optimizer and the categorical cross-entropy loss function. The true negative rate (TNR) is utilized as the performance metric, and early stopping is implemented to prevent model overfitting.

## 3 RESULTS

The input data for the two CNN configurations, ResNet50 and from scratch, is derived from the images obtained through the CWT of the PSD of each signal. Classes $C_0$ and $C_1$ are used for the hyperparameter adjustment stage, distributing 70% of the images for the training data and 30% for the validation data. Class $C_2$, corresponding to the reintubated patients, is used as the test data. Table 2 presents the parameters obtained from the Bayesian optimization algorithm (BOA) for the eight ResNet50 CNNs, along with the AUC index of the ROC curve, which is used to weigh the final prediction probability of the eight ResNet50 CNNs. Table 3 displays four statistical measures for the training and validation datasets: accuracy, recall, precision, and F1 score, for 150 runs.

Table 2. Results of the Bayesian optimization algorithm for CNN ResNet50

| Variable | Neurons of Fully Connected Layers | Optimizer | AUC Index of the ROC Curve |
|---|---|---|---|
| $f/V_T$ | 480 | Adam | 0.9934 |
| $V_T/T_I$ | 64 | Adam | 0.9994 |
| $T_{Tot}$ | 64 | Adam | 1.00 |
| $T_I/T_{Tot}$ | 448 | Adam | 0.9988 |
| $T_I$ | 352 | Adam | 0.9998 |
| $T_E$ | 160 | SGD | 0.9826 |
| $V_T$ | 64 | Adam | 1.00 |
| $RR$ | 192 | SGD | 0.9902 |

Table 3. Statistical Data of the CNN ResNet50

|  | Accuracy | Recall | Precision | F1-Score |
|---|---|---|---|---|
| **Training data** | 97.5 ± 0.5 | 96.8 ± 1.5 | 93.9 ± 1.812 | 95.5 ± 1.751 |
| **Validation data** | 97.8 ± 1.42 | 96.7 ± 3.12 | 94.5 ± 2.26 | 95.6 ± 2.89 |

Table 4 illustrates the architecture of the CNN developed from scratch using the Bayesian optimization algorithm. It commences with a two-dimensional convolution layer (222, 222, 32) with 32 filters at the input. Subsequently, a max-pooling layer is applied, reducing the dimensions to (111, 111, 32). The CNN proceeds with another two-dimensional convolution layer, producing an output matrix of (109, 109, 32), followed by another max-pooling layer that reduces the dimensions to (54, 54, 32). A dropout layer with a node retention probability of 50% is incorporated, followed by a ReLU activation function. The resulting tensor is then flattened into a one-dimensional vector of size 93312. Two fully connected layers, with 1024 and 512 neurons, respectively, are then included before the final SoftMax output layer with 2 neurons, representing the output classes of the model. This design utilizes a hierarchical architecture to progressively process information, extract relevant features, and ultimately classify the input data into the desired categories. The learning rate is set at, $63 \times 10^{-5}$ and the batch size is 32. Table 5 presents the classification system results for both the training and validation data, based on 150 runs.

Table 4. CNN architecture from scratch

| Layer | Output Shape |
|---|---|
| Layer Conv2D | (222, 222, 32) |
| Layer Maxpooling | (111, 111, 32) |
| Layer Conv2D | (109, 109, 32) |
| Layer Maxpoolin | (54, 54, 32) |
| Layer Dropout | (54, 54, 32) |
| Layer ReLU | (54, 54, 32) |
| Layer Flatten | (93312) |
| Layer Fully connected | (1024) |
| Layer Fully connected | (512) |
| Layer Softmax | (2) |

Table 5. Statistical data of the CNN from scratch

|  | Accuracy | Recall | Precision | F1-Score |
|---|---|---|---|---|
| **Training data** | 97.63 ± 0.783 | 97.32 ± 1.713 | 94.65 ± 2.051 | 95.95 ± 1.468 |
| **Validation data** | 98.00 ± 1.821 | 97.24 ± 4.905 | 95.31 ± 5.346 | 96.16 ± 4.060 |

For the test group, class $C_2$, PSD images are loaded, and a prediction is made on the probability of classification using the CNN ResNet50 and CNN from scratch models. The CNN ResNet50 architecture calculates a weighted average using the weights obtained from the ROC of each CNN to obtain the final probability of success or failure for the patient. In all the cases evaluated, the reintubated patients were classified as failures by both CNN architectures. This indicates that, according to the trained models, these patients should not have been extubated after SBT. The consistency in the classification of patients as failures by both architectures suggests a high level of agreement in the predictions made and supports the validity of the approach used for weaning data analysis.

In recent years, there has been progress in the development of ML-based MV weaning prediction models. These models have utilized various tools, such as neural networks, support vector machines, and other ML techniques, to forecast the success of MV weaning. For instance, in a study of 108 patients in a medical intensive care unit of a university hospital, an artificial intelligence prediction model was created, incorporating body mass index (BMI) at admission, occlusion pressure at 0.1 s (P0.1), and heart rate analysis parameters, achieving an overall performance of 62–83% [42]. Additionally, in another study involving 709 patients undergoing pulmonary resection, seven supervised ML algorithms were assessed, and Naive Bayes emerged as the top classifier with an accuracy of 84.5% [43]. Another study, involving 1439 patients from the cardiac ICU of Cheng Hsin General Hospital, gathered 28 variables, including demographic characteristics, arterial gas measurements, and ventilation parameters; a prediction model was developed using support vector machines (ROC-AUC = 88%), logistic regression (ROC-AUC = 86%), and XGBoost (ROC-AUC = 85%) [44]. Similarly, a study that considered demographics, vital signs, laboratory data, and ventilator data to predict extubating failure defined reintubation within 72 hours of extubating. This study employed supervised learning algorithms such as Random Forest, XGBoost, and LightGBM, achieving an accuracy range of 89.7% to and 96% [45].

While the inclusion of multiple parameters may improve the accuracy of artificial intelligence models, it may also complicate their application in clinical practice due to the need to collect and process a large amount of data. Nevertheless, it is essential to consider the full spectrum of respiratory and electrocardiographic flow signals, as changes in patient behavior during SBT may be indicative of their weaning ability [46]. The proposed methodology allows a comprehensive analysis of all data recorded during the SBT test using advanced methods such as CNN, which are able to identify complex patterns in multidimensional data. This provides additional details that traditional ML algorithms might miss, which is crucial to gain a more complete understanding of patient response during weaning tests and improve accuracy in predicting ventilatory weaning success.

A sensitivity analysis of occlusion is conducted to determine the most critical regions of the PSD for classifying a specific class. The method involves iteratively occluding parts of the input image and observing how the classification is impacted by hiding certain regions. The sensitivity of each region is assessed by calculating the difference between the original classification score and the score after occlusion [47]. Figure 5 displays the occlusion analysis image for the ResNet50 CNN related to the $f/V_T$ and $RR$ signals of a patient in a particular class $C_2$. The variation is not greater than ±0.2, suggesting that crucial information for classification is evenly distributed throughout the image. In other words, the CNN extracts various visual features present in the PSD of the CWT without being constrained to specific locations. The process is repeated for the CNN from scratch. The eight images of the CWT are segmented into windows of 40 × 40 pixels and shifted along each dimension with a fixed step of 20 pixels. This iteration is then utilized to assess the sensitivity of each region of the image for the final classification. The resulting image from this occlusion analysis does not exceed the value of ±0.2, indicating that the information required to determine if a patient can be weaned off a ventilator is spread throughout the PSD. This outcome underscores the significance of thoroughly analyzing the information gathered during the 30-minute period covered by the SBT. The pertinent characteristics for classification are not confined to specific regions of the image but are comprehensively distributed throughout the entire spectrogram.

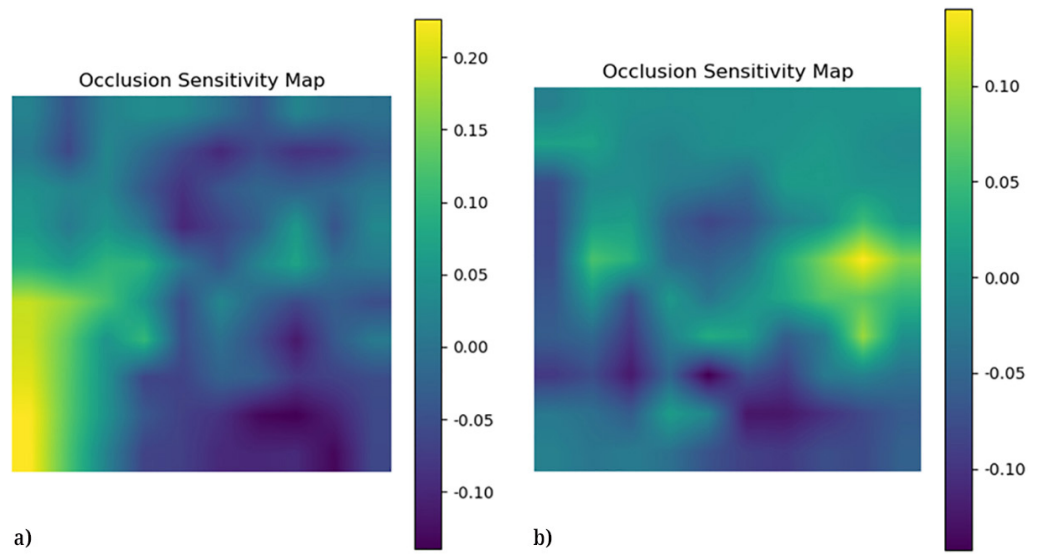

**Fig. 5.** Occlusion sensitivity analysis for a one patient in the Reintubation group.
a) Time series $f/V_T$ b) Time series RR

## 4 DISCUSSION

Developing a patient classification system for individuals at risk of reintubation is essential in the medical field due to the significant risks involved, which can impact morbidity and mortality. The results demonstrate the effectiveness of both ResNet50 and a CNN developed from scratch in classifying patients who have undergone SBT. The tables present statistical data indicating high levels of accuracy, recall, and F1 score in both the training and validation datasets, suggesting that the model performs well in the classification task. Additionally, the consistency in classifying patients as group $C_2$ failures by both CNN architectures indicates a high level of agreement with the predictions made. The analysis of occlusion sensitivity provides detailed information on the critical regions for classification, allowing for a deeper understanding of how the CNN performs the classification task and emphasizing the importance of the information captured in the CWT spectrum for clinical decision-making. CWT decomposes a signal into components that represent different frequency characteristics for various time instants, providing detailed insight into the temporal and frequency structure of the signal. In the context of analyzing the weaning process, the information captured in the CWT serves as a foundation for clinical decision-making.

Performing a SBT evaluates the patient's ability to maintain an adequate breathing pattern without ventilatory support. The response of the respiratory system to this test can provide crucial information about lung function and the patient's capacity to breathe autonomously. On the other hand, the RR signal represents changes in the time interval between successive beats and reflects the cardiovascular system's ability to adapt to physiological demands. During an SBT, RR can indicate the heart's response to spontaneous breathing and offer insights into the patient's cardiovascular stability. The combination of respiratory flow spectra and RR enables a CNN to generate rate and time indicators for assessing the patient's readiness to be weaned from MV. CNN can discern intricate patterns in these multidimensional signals and extract relevant features that signify the patient's ability to maintain stable autonomic breathing. By continuously monitoring these signals during an SBT, CNN can promptly identify signs of deterioration in cardiorespiratory function and make informed decisions regarding patient care. This approach allows for a more precise and sensitive evaluation of the patient's readiness to be weaned from the ventilator, thereby reducing the risk of failed extubating or respiratory complications.

## 5  CONCLUSION

Developing a patient classification system for those at risk of reintubation is crucial in the medical field due to the significant risks associated with the procedure, which can impact morbidity and mortality. The results demonstrate the effectiveness of using CNNs, including both the pre-trained ResNet50 architecture and a CNN from scratch, to classify patients after SBT. The CWT spectrum captures information that provides detailed insight about respiratory and electrocardiographic patterns. This allows accurate discernment of when a patient may need continuous ventilation or can be safely extubated. The methodology used in this study, which combines respiratory flow spectra and heart rate variability, shows promise in assessing a patient's ability to be weaned from the ventilator. Using the real-time information provided by these signals, decisions regarding ventilator weaning in an ICU can be dynamically tailored, allowing a more precise and sensitive evaluation of the patient's ability to breathe independently and steadily. For future research, it is recommended to further investigate the potential of CNNs to classify patients at risk of reintubation, taking into account various architectures and training techniques. Additionally, exploring alternative signal processing and data analysis techniques could enhance the model's accuracy and robustness. Furthermore, conducting more clinical studies is crucial to validating the effectiveness and applicability of this approach in real clinical settings. This will ultimately improve the management and care of patients who undergo extubating processes in the ICU. The development of decision support systems based on artificial intelligence can ultimately have a significant impact on improving clinical outcomes and reducing morbidity and mortality in critically ill patients.

## 7    AUTHORS

**Hernando González** received the B.Eng. degree in electronic engineering from Universidad Industrial de Santander (UIS), the Master in Electronic Engineering from UIS, and the Doctorate in Engineering from UNAB. He is currently an Associate Professor of the Mechatronics Engineering program at UNAB and a member of GICITII. His research interests include machine learning, control systems, signal treatments, robotics, electronics and mechatronics devices. He can be contacted at email: hgonzalez7@unab.edu.co.

**Carlos Julio Arizmendi** received the B.Eng. degree in electronic engineering from Universidad Industrial de Santander (UIS), Bucaramanga, Colombia, in 1997; in 2008, received the diploma of advanced studies in the doctorate of biomedical engineering from UPC; in 2012, he received the PhD in artificial intelligent from UPC. He is currently titular professor of the Mechatronics Engineering program and Biomedical Engineering program at Universidad Autonoma de Bucaramanga (UNAB) is also director of the GICITII research group at UNAB, and is the creator of the Biomedical Engineering Program at UNAB. His research interests include machine learning, deep learning, signal treatments, electronics, mechatronics, and biomedical devices. He can be contacted at email: carizmendi@unab.edu.co.

**Beatriz F. Giraldo** received the B.Eng. degree in electrical engineering from the Technical University of Pereira, Pereira, Colombia, in 1983, the M.Sc. degree in Bioengineering, and the Ph.D. degree in industrial engineering in the biomedical Engineering Program from the Technical University of Catalonia (UPC), Barcelona, Spain, in 1990 and 1996, respectively. She is an associate professor at the Automatic Control Department, UPC, and a senior researcher in the Biomedical Signal Processing and Interpretation (BIOSPIN) group of the Institute for Bioengineering of Catalonia (IBEC) and the CIBER de Bioingeniería, Biomateriales y Nanomedicina (CIBER-BBN) in Spain. Her main research interests include biomedical signal processing and statistical analysis of cardiac, respiratory, and cardiorespiratory signals. The last studies are oriented to enhance knowledge of the respiratory pattern and their interaction with the cardiac system, working with different signals such as the ECG, respiratory flow and volume, and blood pressure. Current research projects include studies of the cardiac and respiratory systems in elderly patients, chronic heart failure patients, and patients in the weaning trial process. She can be contacted at email: Beatriz.Giraldo@upc.edu or bgiraldo@ibecbarcelona.eu.